\documentclass[10pt,conference]{IEEEtran}

\IEEEoverridecommandlockouts \IEEEpubid{\makebox[\columnwidth]{979-8-3315-8951-6/25/\$31.00~\copyright2025 IEEE \hfill} \hspace{\columnsep}\makebox[\columnwidth]{}}

\usepackage{cite}

\ifCLASSINFOpdf
   \usepackage[pdftex]{graphicx}
   \graphicspath{{figs/}}
   \DeclareGraphicsExtensions{.pdf,.jpeg,.png}
\else
   \usepackage[dvips]{graphicx}
   \graphicspath{{../figs/}}
   \DeclareGraphicsExtensions{.eps}
\fi
\usepackage[cmex10]{amsmath}
\usepackage{amsthm}
\usepackage{hyperref}

\usepackage{algorithmic}
\usepackage{tikz}
\usepackage{pdflscape}
\usepackage[margin=0.7in]{geometry}
\usepackage{multicol}
 \usepackage{booktabs}
\definecolor{StudyColor}{RGB}{102,153,255}    
\definecolor{WorkColor}{RGB}{153,204,102}     
\definecolor{CurrentColor}{RGB}{255,153,102}  
 \usetikzlibrary{calc} 

\tikzstyle{descript} = [text=black, align=center, minimum height=1.5cm, outer sep=0pt, font=\footnotesize]
\tikzstyle{activity} = [align=center, outer sep=1pt, font=\scriptsize\bfseries, text=white]

\usepackage{array}

\ifCLASSOPTIONcompsoc
  \usepackage[caption=false,font=normalsize,labelfont=sf,textfont=sf]{subfig}
\else
  \usepackage[caption=false,font=footnotesize]{subfig}
\fi
\usepackage{url}
\usepackage{float}

\usepackage{tabularx}

\newif\iffinal
\finalfalse

\iffinal
\else
\usepackage[switch]{lineno}
\fi

\begin{document}

\title{Sonar Image Datasets: A Comprehensive Survey of Resources, Challenges, and Applications}

\author{
    \IEEEauthorblockN{
         Larissa S. Gomes$^{1}$,
         Gustavo P. Almeida$^{1}$,
         Bryan U. Moreira$^{1}$
         Marco Quiroz$^{1}$,
         Breno Xavier$^{1}$,
         Lucas Soares$^{1}$,\\ 
         Stephanie L. Brião$^{1}$,
         Felipe G. Oliveira$^{1,2}$,
         and Paulo L. J. Drews-Jr$^{1}$ }
    
    \IEEEauthorblockA{
        \textit{$^{1}$Centro de Ciências Computacionais . Universidade Federal do Rio Grande}, Rio Grande, RS, Brasil \\
        Email: \{larissaesilva,
        gustavo.pereira.furg,
        bryan,
        mquiroz,
        breno.paula,
        lucas.soares,
        stephanie.loi,
        paulodrews\}@furg.br
    }
    \IEEEauthorblockA{
        \textit{$^{2}$Instituto de Ciências Exatas e Tecnologia (ICET). Universidade Federal do Amazonas}, \\Itacoatiara, AM, Brasil\\
        Email: felipeoliveira@ufam.edu.br
    } 
}

\maketitle

\begin{abstract}
Sonar images are relevant for advancing underwater exploration, autonomous navigation, and ecosystem monitoring. However, the progress depends on data availability. The scarcity of publicly available, well-annotated sonar image datasets creates a significant bottleneck for the development of robust machine learning models. This paper presents a comprehensive and concise review of the current landscape of sonar image datasets, seeking not only to catalog existing resources but also to contextualize them, identify gaps, and provide a clear roadmap, serving as a base guide for researchers of any kind who wish to start or advance in the field of underwater acoustic data analysis. We mapped publicly accessible datasets across various sonar modalities, including Side Scan Sonar (SSS), Forward-Looking Sonar (FLS), Synthetic Aperture Sonar (SAS), Multibeam Echo Sounder (MBES), and Dual-Frequency Identification Sonar (DIDSON). An analysis was conducted on applications such as classification, detection, segmentation, and 3D reconstruction. This work focuses on state-of-the-art advancements, incorporating newly released datasets. The findings are synthesized into a master table and a chronological timeline, offering a clear and accessible comparison of characteristics, sizes, and annotation details datasets. 

\end{abstract}

\IEEEpeerreviewmaketitle


\section{Introduction}

Underwater perception is a fundamental frontier not just for robotics, but also for the entire field of oceanographic research, ecosystem monitoring, analysis, and security purposes~\cite{drews2016underwater}. The exploration of the underwater environment is an admittedly challenging task due to the severely limited effectiveness of multiple sensors, such as GPS (Global Positioning System)~\cite{dos2020matching} and optical sensors~\cite{schein2024udbe}. Besides the absorption of radio-like signals, the water causes light absorption, scattering, and turbidity~\cite{mello2022underwater}, which makes cameras and LiDAR (Light Detection and Ranging) ineffective for long-range observation~\cite{Fallon}. In this context, sonar (Sound Navigation and Ranging) emerges as a fundamental perception technology, as its sound waves, which can propagate efficiently underwater, can \textit{see} where light cannot reach~\cite{machado}. 

Technology has applied sonar in a vast range of applications. The advancement of Artificial Intelligence (AI), particularly deep learning, and the resulting enhanced ability to automatically analyze sonar data for tasks such as object detection, seafloor mapping, and autonomous navigation, regardless of visibility conditions~\cite{Williams2022}, has established sonar as the cornerstone of modern underwater robotics and exploration.


Sonar is not a monolithic term; it encompasses a variety of modalities, such as Side-Scan Sonar (SSS), ideal for mapping large areas~\cite{SSS}, Forward-Looking Sonar (FLS), crucial for the navigation of autonomous vehicles~\cite{Fallon}, Synthetic Aperture Sonar (SAS), which offers very high-resolution images and Multibeam Echosounder (MBES), that can simultaneously collect data across a wide area~\cite{Brown2011}. At the same time, Dual-Frequency Identification Sonar (DIDSON) can produce near-video-quality imagery~\cite{zhu2024method}.

Each type of sonar generates data with different characteristics, resolutions, and artifacts, making expertise in one data type difficult to transfer to another~\cite{Williams2022}. A later section of this article will detail these modalities, providing a technical basis for understanding the heterogeneity of the available datasets.



This article aims to provide a comprehensive and concise review of the current landscape of sonar datasets, seeking not only to catalog existing resources but also to contextualize them, identify gaps, and provide a clear roadmap, serving as an essential guide for researchers of any kind who wish to start or advance in the field of underwater acoustic data analysis. Compared to previous reviews, this work stands out for incorporating the latest datasets and their applications that have emerged in recent years, thereby reflecting the current state of the art. The main contributions of this work are summarized as follows:
\begin{itemize}
    \item We present a clear and simplified overview of five main sonar types, providing essential context for researchers new to acoustic imaging.
    \item We compile and map publicly available sonar image datasets published in the last five years, highlighting their characteristics and accessibility.
    \item We provide a detailed discussion of typical applications and challenges associated with sonar images, including noise, resolution, and domain-specific constraints.
\end{itemize}



\section{Related Work}

The rapid development of deep learning has driven significant progress in sonar-based perception for underwater environments. Surveys focusing on automatic target recognition (ATR) form one of the earliest areas of research. Neupane et al.~\cite{neupane2020review} enumerate existing datasets, outline key deep learning techniques, and highlight challenges such as data scarcity and variability in underwater conditions.

Steiniger et al.~\cite{steiniger2022survey} expand this discussion by providing a comprehensive survey on deep learning-based computer vision for sonar imagery, with a particular focus on ATR. Unlike Neupane et al.~\cite{neupane2020review}, they do not systematically enumerate the available sonar datasets; instead, they emphasize the potential of synthetic aperture sonar (SAS) and the need for robust feature extraction methods in varying acoustic conditions.

Khan et al.~\cite{khan2024underwater} present an updated perspective on underwater target detection using deep learning, discussing practical applications and methodological trends. However, their survey lacks an in-depth discussion of available datasets, focusing instead on detection pipelines and challenges such as generalization across domains.

Other works emphasized segmentation and recognition frameworks. Teng and Zhao~\cite{teng2020underwater} similarly analyze underwater target recognition frameworks within the deep learning paradigm. Their work outlines the performance of different architectures. Still, it omits a detailed inventory of sonar datasets, highlighting instead the need for adaptable frameworks that can integrate multi-sensor data, including color images and sonar.

Tian et al.~\cite{tian2020review} contribute a more specialized survey dedicated to wavelet methods for sonar image segmentation. This work focuses exclusively on segmentation tasks and does not consider the broader scope of classification and detection applications.

A third line of surveys addressed application-specific domains. In the field of biological monitoring, Yassir et al.~\cite{yassir2023acoustic} systematically review the use of deep learning for acoustic fish species identification. Although their review addresses different sensor types, including multibeam echosounders (MBES), it does not provide a detailed enumeration of sonar datasets. It focuses on classification and segmentation tasks relevant to the fisheries industry.

Chai et al.~\cite{chai2023deep} also examine deep learning applications for sonar imagery in aquaculture, providing a more extensive enumeration of datasets. They discuss challenges in object detection and segmentation within underwater environments, emphasizing how dataset availability shapes algorithm development.


Finally, broader reviews have explored multimodal and general underwater perception tasks. More recently, Aubard et al.~\cite{aubard2025sonar} provide a broad overview of sonar-based deep learning in underwater robotics. They discuss robustness and practical deployment challenges, enumerating datasets and highlighting the need for SLAM integration with sonar-based perception pipelines.

Zhang et al.~\cite{ZHANG2025121862} focus on sonar image enhancement, addressing noise reduction techniques to improve object recognition accuracy. While they thoroughly review denoising methods and domain adaptation for underwater scenes, they do not present a comprehensive dataset mapping.

Despite these valuable contributions, existing surveys tend to focus on specific tasks, limited sonar modalities, or narrow application areas. Our work aims to bridge these gaps by providing an extensive mapping and technical analysis of acoustic image datasets across SSS, SAS, MBES, FLS, and DIDSON systems. We also explore the most recent and relevant databases, detailing their applications, technical characteristics, and limitations. Ultimately, our goal is to deliver a comprehensive and practical review that clarifies how these datasets can support various sonar-based deep learning tasks, highlighting current trends, challenges, and directions for future research.

\begin{figure*}[htpb]
    \centering

    \subfloat[SSS\cite{sethuraman2025machine}\label{fig:sss}]{
        \includegraphics[width=0.15\textwidth, height=0.15\textwidth, keepaspectratio=false]{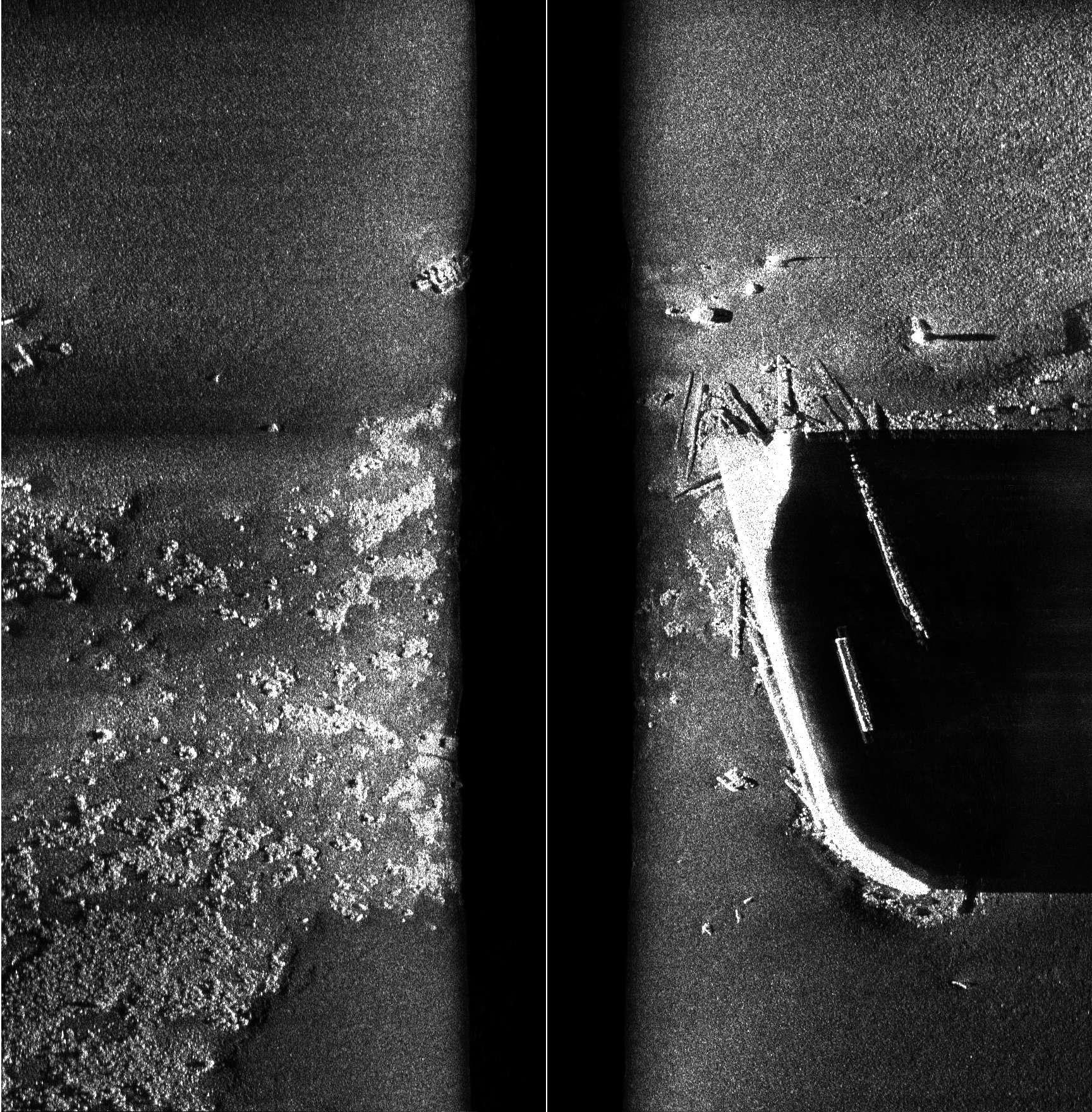}
    }
    \hfill
    \subfloat[FLS\cite{9607740}\label{fig:fls}]{
        \includegraphics[width=0.15\textwidth, height=0.15\textwidth, keepaspectratio=false]{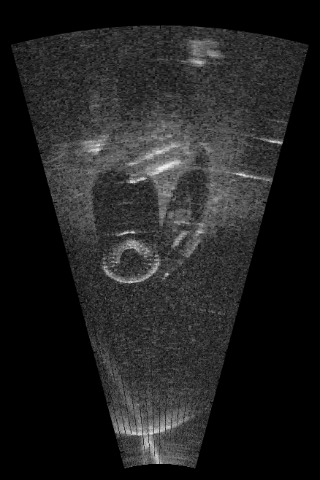}
    }
    \hfill
    \subfloat[SAS\cite{cobb2022sassed}\label{fig:sas}]{
        \includegraphics[width=0.15\textwidth, height=0.15\textwidth, keepaspectratio=false]{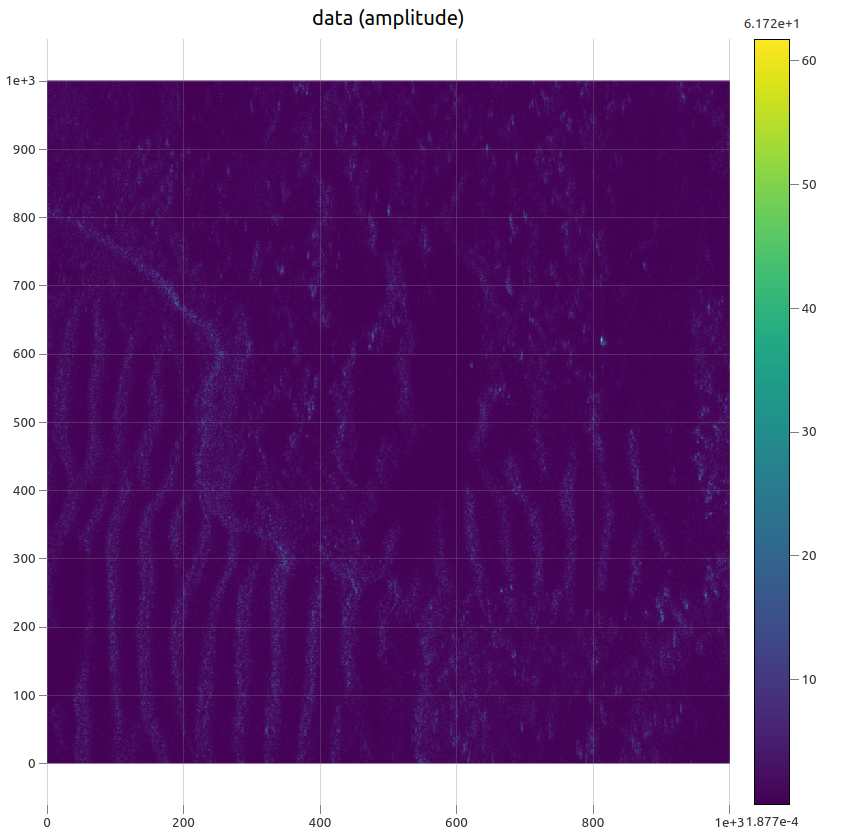}
    }
    \hfill
    \subfloat[MBES\cite{multibeam_bathymetry_UK}\label{fig:mbes}]{
        \includegraphics[width=0.15\textwidth, height=0.15\textwidth, keepaspectratio=true]{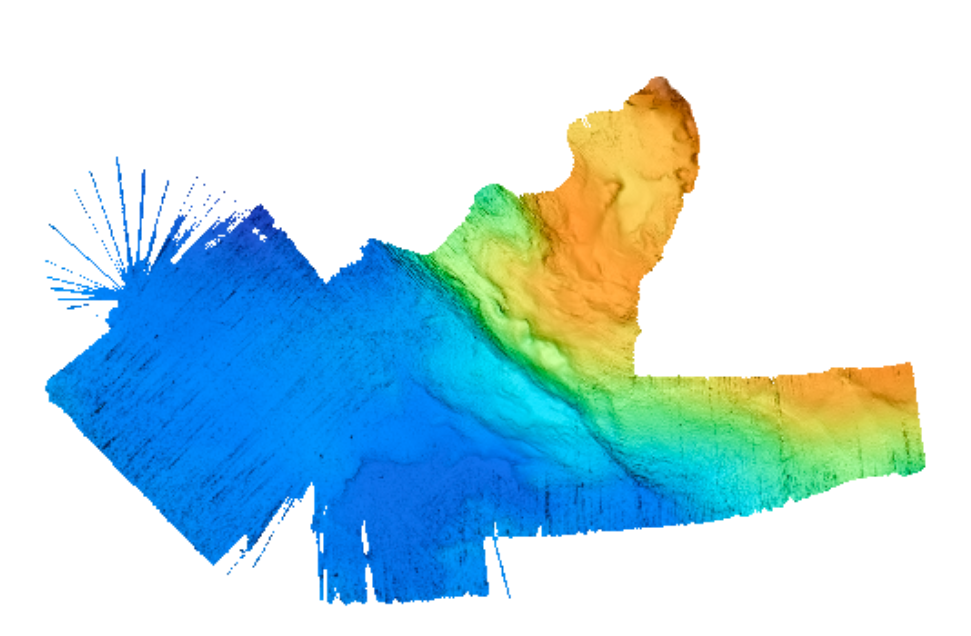}
    }
    \hfill
    \subfloat[DIDSON\cite{wehbe2022sonar}\label{fig:didson}]{
        \includegraphics[width=0.15\textwidth, height=0.15\textwidth, keepaspectratio=true]{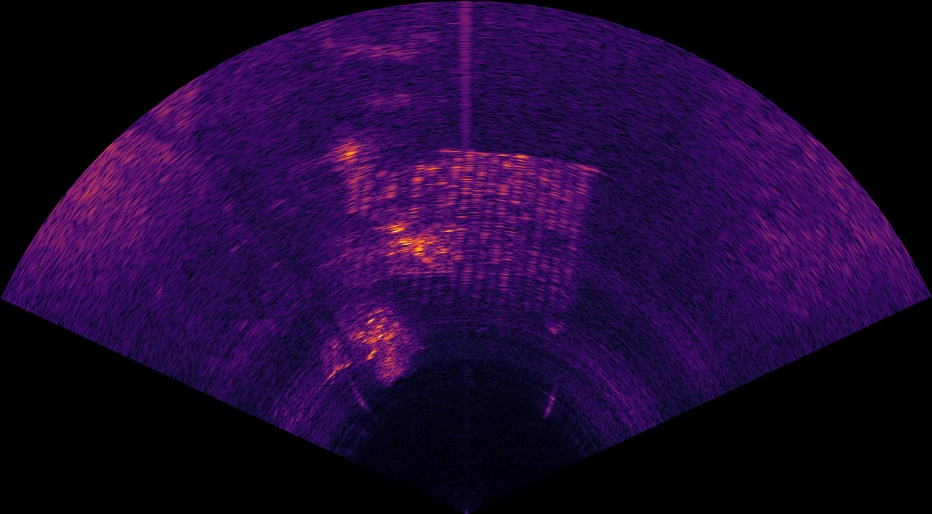}
    }

    \caption{Sonar images: (a) Side Scan Sonar, (b) Forward Looking Sonar, (c) Synthetic Aperture Sonar, (d) Multibeam Echo Sounder, and (e) DIDSON.}
    \label{fig:sonar_matrix}
\end{figure*}

\section{Sonar Types and their Features}

This section provides an overview of the five main sonar types employed across various underwater imaging datasets. Each sonar modality offers distinct capabilities in data acquisition and is suitable for different environmental and operational conditions. The classification is based on the dataset names and associated metadata. 

\textbf{Side-Scan Sonar} (SSS) operates based on fan-shaped acoustic beams, narrow in the along-track dimension and wide in the vertical dimension~\cite{brenden2024evolution}. 
The data, however, is significantly susceptible to noise and environmental factors. This limits its usage in quantitative applications, as machine learning for automated target recognition is often required to overcome high false alarm rates~\cite{li2024side}.
Fig. \ref{fig:sss} shows a representation of data collected by an SSS sonar.



\textbf{Forward-Looking Sonar} (FLS) is an active acoustic sensor which works by emitting a fan of acoustic beams in the forward direction, generating a 2D or 3D image of obstacles and targets ahead~\cite{franchi20202d}.
It is often used for long-range tracking of underwater targets, as optical systems are often compromised by turbidity and low-light conditions.~\cite{cho2023advanced}. Otherwise, the most significant challenges for FLS technology involve high signal processing required~\cite{cho2023advanced}.
See Fig. \ref{fig:fls} to visualize an image produced by this type of sonar.



\textbf{Synthetic Aperture Sonar} (SAS) is an imaging technique that can achieve a well-balanced spot between range and resolution, something rare in conventional sonar systems~\cite{hansen2019synthetic, sternlicht2015historical}. It works by combining the acoustic returns from multiple successive pings along the movement of the sonar platform~\cite{wagner2020synthetic}.
An instance of imagery produced by SAS sonar is in Fig. \ref{fig:sas}.


A \textbf{Multibeam Echosounder} (MBES) is the standard hydrographic tool for high-resolution seafloor mapping, acquiring bathymetric data across a wide swath perpendicular to a vessel's track with a single acoustic transmission~\cite{dosits2019multibeam}. 
The accuracy of all MBES data is critically dependent on robust compensation for vessel motion and precise measurement of the water column's sound velocity profile~\cite{lurton2002introduction}.
Fig. \ref{fig:mbes} is an example of an image produced by MBES sonar.


\textbf{ Dual-Frequency Identification Sonar} (DIDSON) is a high-frequency, multibeam imaging sonar, often referred to as an "acoustic camera"~\cite{ belcher2002dual}. It operates based on a set of acoustic lenses to focus sound energy onto a transducer array, producing near-video-quality imagery with exceptionally high resolution~\cite{zhu2024method}.
DIDSON images usually take the form as in Fig.~\ref{fig:didson}. 

Across these sonar modalities, acoustic datasets differ not only in resolution and image geometry but also in how labels are represented. SSS and SAS images often resemble standard grayscale textures commonly used in classification networks, whereas FLS is designed for real-time segmentation. MBES requires geometric or topographic analysis tools. The diversity of sensors, annotation styles, and operational conditions across these datasets presents unique opportunities and challenges for sonar-based perception systems.


\section{Overview of Sonar Image Datasets}

This section reviews publicly available sonar datasets that cover the main sonar modalities discussed previously, supporting tasks ranging from classification and detection to segmentation, reconstruction, SLAM, tracking, and image translation. 

Table~\ref{tab:sonar-datasets} organizes the reviewed datasets by sonar modality (e.g., Side-Scan, FLS, MBES) and by their primary application task (classification, detection, segmentation, etc.). This categorization provides a clearer structure for identifying datasets suited to specific research goals.

\begin{figure}[htp]
\centering
\resizebox{0.48\textwidth}{!}{%
\begin{tikzpicture}[very thick, scale=1, every node/.style={transform shape}]
    \footnotesize

    \definecolor{mbescol}{rgb}{1.0,0.5,0.0}   
    \definecolor{sascol}{rgb}{0.6,0.0,0.6}   

    \tikzset{sonarline/.style={->, very thick}}

    \coordinate (Start) at (0,0);
    \coordinate (End)   at (21,0);

    \fill[green!30] (0,0.0) rectangle (12.5,-0.8);
    \node[align=center] at (5,-0.65) {\tiny Early Datasets};

    \fill[yellow!60!red, fill opacity=0.30] (12.5,0.0) rectangle (21,-0.8);
    \node[align=center] at (14,-0.65) {\tiny New Datasets};

    \draw[->, thick] (Start) -- (End);
    \foreach \x/\year in {0/2020, 2/2021, 4/2022, 8.2/2023, 12.5/2024, 17/2025} {
        \draw (\x,3pt) -- (\x,-3pt) node[below=4pt] {\year};
    }
    \draw (20,0.2) node[right] {\textit{Future}};

    \foreach \x/\name/\dy/\col in {
        0.5/Seabed\\Objects/1.0/blue,
        1.2/NNSSS/1.8/blue,
        2.2/DIDSON/1.0/red,
        2.9/SCTD/1.8/blue,
        3.6/FLS Marine\\Debris/2.5/green,
        4.9/UATD/1.8/green,
        5.5/MBES\\SLAM/2.5/mbescol,
        6.3/Arctic Cruise\\MSM95/3.5/mbescol,
        7.2/FLS\\Detection/4.5/green,
        7.9/Sonar\mbox{-}to\mbox{-}RGB\\Translation/5.5/red,
        8.7/Marine\\PULSE/1.0/blue,
        9.3/SASSED/1.8/sascol,
        10.0/Seafloor\\Sediments/2.5/blue,
        11.6/MSM30/4.2/mbescol,
        12.2/Camera Sonar\\Fusion/4.6/red,
        13.3/NKSID/5.5/green,
        14.0/SSS Mine\\Detection/4.5/blue,
        14.6/SWDD/4.0/blue,
        15.3/SubPipe/3.5/blue,
        15.8/UXO/2.5/green,
        16.2/SAS/1.8/sascol,
        17.5/AI4Shipwreck/1.0/blue,
        18.5/SEE /2.0/green,
        19.2/MBES\\UK EA/3.0/mbescol}{
        \draw[sonarline, draw=\col] (\x,0) -- ++(0,\dy);
        \fill[\col] (\x,0) circle (2pt);
        \node[align=center, text width=2.8cm, anchor=south] at (\x,\dy) {\footnotesize \name};
    }

    \def\xA{4.4}\def\dyA{1.0}
    \draw[sonarline, draw=mbescol, line width=1.6pt] (\xA-0.06,0) -- ++(0,\dyA);
    \draw[sonarline, draw=blue,    line width=1.6pt] (\xA+0.06,0) -- ++(0,\dyA);
    \fill[mbescol] (\xA-0.06,0) circle (2pt);
    \fill[blue]    (\xA+0.06,0) circle (2pt);
    \node[align=center, text width=2.8cm, anchor=south] at (\xA,\dyA) {\footnotesize Aurora};

    \def\xS{10.7}\def\dyS{3.5}
    \draw[sonarline, draw=green,   line width=1.6pt] (\xS-0.06,0) -- ++(0,\dyS);
    \draw[sonarline, draw=mbescol, line width=1.6pt] (\xS+0.06,0) -- ++(0,\dyS);
    \fill[green]   (\xS-0.06,0) circle (2pt);
    \fill[mbescol] (\xS+0.06,0) circle (2pt);
    \node[align=center, text width=2.8cm, anchor=south] at (\xS,\dyS) {\footnotesize Shipwreck};

    \begin{scope}[shift={(13.3,7.3)}]
        \draw[rounded corners, thin] (-0.3,-0.55) rectangle (7.0,0.75);
        
        \draw[sonarline, draw=blue]    (0,0.5) -- +(0.5,0); 
        \node[anchor=west] at (0.65,0.5)  {\scriptsize \textbf{SSS} (Side-Scan)};
        
        \draw[sonarline, draw=green]   (0,0.1) -- +(0.5,0); 
        \node[anchor=west] at (0.65,0.1)  {\scriptsize \textbf{FLS} (Forward-Looking)};
        
        \draw[sonarline, draw=sascol]  (3.3,0.5) -- +(0.5,0); 
        \node[anchor=west] at (3.95,0.5) {\scriptsize \textbf{SAS} (Synthetic Aperture)};
        
        \draw[sonarline, draw=mbescol] (3.3,0.1) -- +(0.5,0); 
        \node[anchor=west] at (3.95,0.1) {\scriptsize \textbf{MBES} (Multibeam Echo)};
        
        \draw[sonarline, draw=red]     (0,-0.3) -- +(0.5,0); 
        \node[anchor=west] at (0.65,-0.3) {\scriptsize \textbf{DIDSON}};
    \end{scope}

\end{tikzpicture}}
\caption{\textbf{Timeline of sonar dataset publications (2020–2025).} Colorful lines/arrows by modality:
\textbf{SSS} (blue), \textbf{FLS} (green), \textbf{SAS} (purple), \textbf{MBES} (orange), \textbf{DIDSON} (red). Hybrid cases appear with \emph{two parallel arrows}.}
\label{fig:sonar_timeline}
\end{figure}
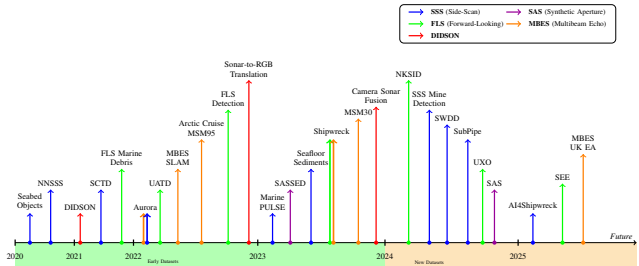

Beyond providing a temporal overview, Fig.~\ref{fig:sonar_timeline} also reveals notable trends in the evolution of sonar datasets. There has been a clear increase in datasets dedicated to segmentation tasks, reflecting the community’s shift toward more detailed scene understanding.

Several notable datasets employ Side‐Scan Sonar (SSS) imagery for object recognition and seabed analysis. The SeabedObjects dataset~\cite{huo2020underwater} includes shipwreck and airplane wreck “waterfall” images captured with towed systems from manufacturers such as L-3 Klein and EdgeTech. Marine PULSE~\cite{du2023revealing} adds annotated examples of pipelines, cables, mounds, and platforms, while the SSS Mine Detection dataset~\cite{santos2024side} provides labeled samples of mine-like objects from Teledyne Gavia AUV surveys. Infrastructure-oriented datasets, such as SWDD~\cite{aubard2024rosar} and SubPipe~\cite{alvarez2024subpipe}, focus on detecting walls and pipelines, respectively, with high-quality manual annotations and pose data for SLAM validation. The NNSSS benchmark~\cite{burguera2020line} and Seafloor Sediments~\cite{rajani2023convolutional} target fine-grained seabed segmentation. Complementing these, SASSED~\cite{cobb2022sassed} delivers high-frequency synthetic aperture sonar snippets for studying seafloor texture mapping, while AI4Shipwreck~\cite{sethuraman2025machine} concentrates on shipwreck segmentation and anomaly detection.

Forward-Looking Sonar (FLS) datasets contribute varied benchmarks for detection and segmentation under realistic acoustic clutter. NKSID~\cite{jiao2024open} covers multiple object classes for open-set recognition, whereas UATD~\cite{xie2022dataset} focuses on tires, mannequins, and boxes with bounding-box labels. FLSMarine~\cite{du2023revealing} and the UXO dataset~\cite{dahn2024acoustic} expand segmentation and detection to debris and unexploded ordnance scenarios. FLSDetection~\cite{Zhu2023ForwardLookingSonarDetectionDataset} provides additional search-and-rescue examples, and SEE~\cite{de2025synthetic} generates synthetic echoes of geometric targets for evaluating reconstruction methods. The DIDSON Fish dataset~\cite{perivolioti2021optimising}, which utilizes a high-frequency DIDSON unit, focuses on recognizing and counting fish species in challenging natural river conditions.

Multibeam Echo Sounder (MBES) datasets emphasize precise bathymetric mapping and SLAM evaluation. The AURORA dataset~\cite{bernardi2022aurora} and MBES-SLAM~\cite{krasnosky2022bathymetric} include synchronized navigation data and seafloor maps for testing localization accuracy. Arctic Cruise MSM95~\cite{dreutter2022mbpd} and the UK Environment Agency MBES dataset~\cite{multibeam_bathymetry_UK} provide real-world seafloor data for monitoring and coastal analysis. The Shipwreck dataset~\cite{pecheux2023self} combines FLS and MBES to support 3D reconstructions and sensor calibration tasks. These collections highlight typical MBES challenges, including navigation drift and gaps in coverage due to acoustic dropouts.

Additional resources explore cross-modal applications: the Sonar-to-RGB Translation dataset~\cite{wehbe2022sonar} and Camera Sonar Fusion~\cite{koken_2023_10220989} pair sonar with optical data to train models for diver tracking and multi-sensor reconstruction in poor visibility. The SCTD dataset~\cite{zhang2021self} complements detection research with varied common targets, and the SAS dataset~\cite{blanford2024air} supports controlled experiments on target scattering in complex environments. Collectively, these datasets—publicly available under permissive licenses—provide a valuable foundation for developing and benchmarking sonar-specific algorithms in underwater robotics and perception research.

\begin{table}[htbp]\tiny
\centering
\renewcommand{\arraystretch}{0.9}
\caption{Sonar Image Datasets: Categorization by Modality and Task. Acronyms: C (Classification), S (Segmentation), D (Detection), R (3D Reconstruction), SL (Simultaneous Localization and Mapping), IT (Image Translation), and T (Tracking).}
\label{tab:sonar-datasets-categorized}
\label{tab:sonar-datasets}
\begin{tabularx}{0.48\textwidth}{
  >{\raggedright\arraybackslash}p{2.1cm} 
  >{\raggedright\arraybackslash}p{0.9cm} 
  >{\raggedright\arraybackslash}p{1.1cm} 
  >{\raggedright\arraybackslash}X 
  >{\raggedright\arraybackslash}p{0.6cm}
}
\toprule
\textbf{Dataset} & \textbf{Sonar} & \textbf{Nº Data} & \textbf{Object Labels} & \textbf{Task} \\
\midrule
\multicolumn{5}{c}{\textbf{Side-Scan Sonar (SSS)}} \\
\midrule
SeabedObjects \cite{huo2020underwater} & SSS & 1,190 & Wrecks/Humans/Mines & C \\
Marine PULSE \cite{du2023revealing} & SSS & 627 & Pipes/Mounds/Platforms & C \\
SSS Mine Detection \cite{santos2024side} & SSS & 1,170 & Mines & D \\
SWDD \cite{aubard2024rosar} & SSS & 7,904 & Walls & D \\
SubPipe \cite{alvarez2024subpipe} & SSS & 10,030 & Pipelines & D \\
NNSSS \cite{burguera2020line} & SSS & 10 & Sea grass/Rocks/Sand & S \\
Seafloor Sediments \cite{rajani2023convolutional} & SSS & 434,164 & Rocks/Marine life & S \\
AI4Shipwreck \cite{sethuraman2025machine} & SSS & 286 & Shipwreck/Non-shipwreck & S,R \\
SCTD \cite{zhang2021self} & SSS & 357 & Aircraft/Ship & D \\
\midrule
\multicolumn{5}{c}{\textbf{Forward-Looking Sonar (FLS)}} \\
\midrule
NKSID \cite{jiao2024open} & FLS & 2,617 & Infrastructures/Propellers/Tires & C \\
UATD \cite{xie2022dataset} & FLS & 9,200 & Tires/Mannequins/Boxes & D \\
UXO \cite{dahn2024acoustic} & FLS & 74,437 & Unexploded ordnance & D,R \\
FLS Marine \cite{du2023revealing} & FLS & 2,471 & Infrastructure/Debris & S \\
FLS Detection \cite{Zhu2023ForwardLookingSonarDetectionDataset} & FLS & 3,752 & Victim/Boat/Plane & D \\
SEE \cite{de2025synthetic} & FLS & 8,454 images, 4,667 sonar & Solids/Helices/Materials & R \\
\midrule
\multicolumn{5}{c}{\textbf{Synthetic Aperture Sonar (SAS)}} \\
\midrule
SASSED \cite{cobb2022sassed} & SAS & 129 & Mud/Sea grass/Rocks/Sand & S \\
SAS \cite{blanford2024air} & SAS & 611 & Targets/Backgrounds & C,D \\
\midrule
\multicolumn{5}{c}{\textbf{Multibeam Echo Sounder (MBES)}} \\
\midrule
Aurora \cite{bernardi2022aurora} & MBES, SSS & MBES: 81km, SSS: 15h & Seabed/Marine habitats & SL \\
MBES-SLAM \cite{krasnosky2022bathymetric} & MBES & 4 missions & Seabed & SL \\
Shipwreck \cite{pecheux2023self} & FLS/MBES & 17,572(mono), 8,577(sonar) & Shipwreck & R \\
Arctic Cruise MSM95 \cite{dreutter2022mbpd} & MBES & 450 & Seafloor & R \\
MSM30 \cite{hanebuth2023mbrd} & MBES & 7,374 & Seafloor & R \\
MBES UK EA \cite{multibeam_bathymetry_UK} & MBES & 26 & Seafloor & R \\
\midrule
\multicolumn{5}{c}{\textbf{Dual-Frequency Identification Sonar (DIDSON)}} \\
\midrule
DIDSON \cite{perivolioti2021optimising} & DIDSON & 1,000 & Fish species & S \\
Sonar-to-RGB \cite{wehbe2022sonar} & DIDSON & 11,810 & Diver & IT \\
Camera Sonar Fusion \cite{koken_2023_10220989} & DIDSON & 1,193 & Diver & D,T \\
\bottomrule
\end{tabularx}
\end{table}


\section{Challenges and Applications}

Working with sonar imagery presents distinct challenges, including complex acoustic propagation effects such as multipath reflections, acoustic shadowing, speckle noise, and frequency-dependent signal attenuation, which vary across different sonar modalities. For example, SSS imagery often exhibits geometric distortions, heterogeneous backscatter, and pronounced acoustic shadows behind raised objects, all of which complicate the extraction of robust features and the generalization of these features to new survey sites.


FLS and DIDSON data images typically suffer from low contrast, pervasive speckle noise, limited dynamic range, and non-uniform illumination. Small objects of interest may be partially occluded or exhibit significant pose variation, further complicating robust detection and classification.


A major practical obstacle is developing reliable automatic detection pipelines in cluttered and dynamic acoustic scenes. False positives can be frequent due to reverberation artifacts, marine debris, and variable seafloor textures that closely resemble objects of interest. Designing robust detectors demands large volumes of diverse, high-quality training data that reflect real-world variability—yet such datasets remain costly and time-consuming to collect and annotate.

For 3D reconstruction tasks, sonar-specific limitations further complicate the process of accurate mapping. Acoustic shadows, decreasing resolution with distance, and pose uncertainties degrade point cloud quality and introduce gaps in reconstructed surfaces. Technologies such as MBES and SAS help mitigate some of these issues but require precise navigation and sophisticated processing pipelines. Sensor fusion with optical or inertial systems can reduce drift and ambiguity. Still, it also presents challenges in calibration and alignment, especially in turbid or low-visibility conditions where cameras tend to underperform.

The lack of standardized benchmarking protocols and comprehensive public datasets for tasks like ATR and large-scale 3D mapping limits the fair comparison of algorithmic advances. Addressing these challenges requires robust signal processing, domain adaptation, the generation of realistic synthetic data, and the development of accessible benchmarks tailored to sonar-specific scenarios.

In the field of object classification, the work presented in~\cite{jmse13030424} uses the Marine-PULSE dataset to recognize and differentiate among four classes of underwater engineering structures in SSS imagery. Employing a  CNN (convolutional neural network) based on the GoogleNet architecture with transfer learning, the authors achieved classification accuracy of over 92\%. Additionally, they demonstrated the effectiveness of data augmentation techniques in improving model performance under data scarcity. These results are highly relevant for applications such as subsea infrastructure inspection, seabed mapping, and automated anomaly detection, where manual interpretation is traditionally demanding and costly.

For object detection, the study in~\cite{jmse12122326} proposes the MFF-YOLOv7 model, which enhances the YOLOv7 detection framework with multi-scale feature fusion and attention mechanisms, significantly improving underwater target recognition. Evaluated on datasets including UATD, the model demonstrated superior detection performance in noisy and cluttered environments. These advancements are crucial for applications in underwater infrastructure inspection and deployment, marine surveillance, and the localization of man-made structures or hazards, where real-time and accurate detection is essential.

In another detection-related application, the work in~\cite{electronics13193874} presents an innovative architecture for sonar image processing, focusing on small-target detection in SSS imagery. The proposed method combines shallow, robust downsampling and dynamic upsampling techniques to retain feature information that is lost in traditional convolutional backbones. The experiments were conducted on the AI4Shipwrecks dataset, where the model achieved a 4.4\% improvement in the mAP50 dataset over the baseline, supporting shipwreck detection and high-resolution structure localization. These findings suggest promising use cases in underwater archaeology and search-and-rescue missions.

Still within the realm of detection, the work in~\cite{wang2025enhancingobjectdetectionaccuracy} investigates the impact of applying deep learning-based denoising methods to sonar images before object detection. Using the Forward-Looking Sonar Detection Dataset, the authors evaluated multiple denoising models originally developed for optical images. The results demonstrated that denoised sonar imagery significantly enhances the accuracy of downstream detection models. This approach is particularly valuable in underwater robotics, where sensor noise can severely degrade the reliability of perception systems operating in dynamic or low-visibility environments.

For segmentation tasks, the method introduced in~\cite{fishes9090346} combines the DIDSON dataset with YOLOv5 and DeepSort to detect and count fish species in real-world reservoir environments. The pipeline eliminates manual labeling and counting by providing automatic detection and multi-object tracking across video frames. With accuracies exceeding 83\%, this work demonstrates practical value for fisheries management and biodiversity monitoring, offering a scalable and efficient alternative to traditional sonar processing tools, such as Echoview.

In the area of SLAM (Simultaneous Localization and Mapping) and geospatial reconstruction, the work in~\cite{rs16071163} conducts a comprehensive survey of seabed mapping techniques and highlights the utility of datasets such as Aurora for autonomous underwater navigation. This dataset, comprising MBES and SSS recordings collected from AUV (Autonomous Underwater Vehicle) missions, enables the simultaneous localization and mapping of complex seafloor structures. The analysis highlights how deep learning and multimodal fusion are transforming underwater geophysics and habitat modeling, while also identifying key challenges in integrating geometric consistency, sensor alignment, and large-scale scene reconstruction in real-world deployments.

In summary, by mapping these datasets, clarifying their typical challenges, and connecting them to emerging applications, this review aims to provide a reference for future research. Addressing the domain-specific limitations highlighted here is key to advancing perception pipelines and the practical deployment of sonar image-based systems in the underwater domain.

\section{Conclusions}

This paper provides a comprehensive and up-to-date review of the sonar image ecosystem, highlighting public datasets in advancing underwater robotics and its applications. We systematically mapped available datasets, including the types: Side Scan Sonar (SSS), Forward Looking Sonar (FLS), Synthetic Aperture Sonar (SAS), Multibeam Echo Sounder (MBES), and Dual-Frequency Identification Sonar (DIDSON). Furthermore, an analysis was performed to explore the use of sonar images in tasks such as classification, detection, segmentation, and 3D reconstruction. Our findings are presented in a table and a chronological timeline, which provide a comparison of dataset characteristics and highlight the growth in data availability over the past five years. By cataloging existing resources and identifying current gaps, this work serves as a base guide for the community, providing a clear roadmap not only to navigate but also to contribute to the field of underwater acoustic image analysis. However, data heterogeneity between different sonar types, along with the influence of acoustic noise, remains a limitation for developing robust learning models. 

It is important to note that this survey does not aim to benchmark dataset quality, as crucial factors such as sensor calibration, resolution, and environmental consistency are often unavailable. Instead, our main contribution lies in mapping, categorizing, and consolidating existing resources, providing the community with a structured and up-to-date reference.


\section*{Acknowledgment}
We want to thank the Coordination for the Improvement of Higher Education Personnel (CAPES) and the Research Support Foundation of the State of Rio Grande do Sul (FAPERGS) for their financial and institutional support. This work was also partly supported by CNPq, FINEP, and Human Resources Program of the National Agency of Petroleum, Natural Gas, and Biofuels (PRH/ANP–PRH22.1/FURG).

%



\end{document}